\documentclass[utf8]{FrontiersinHarvard} 

\usepackage{url,hyperref,lineno,microtype,subcaption}

\usepackage[onehalfspacing]{setspace}

\def\keyFont{\fontsize{8}{11}\helveticabold }
\def\firstAuthorLast{Mubarak {et~al.}} 
\def\Authors{Hamdy Mubarak\,$^{1,*}$, Samir Abdaljalil\,$^{1}$, Azza Nassar\,$^{2}$ and Firoj Alam\,$^{1}$}
\def\Address{$^{1}$Qatar Computing Research Institute, Hamad Bin Khalifa University, Doha, Qatar \\
$^{2}$Hamad Bin Khalifa University, Doha, Qatar  }
\def\corrAuthor{Corresponding Author}

\def\corrEmail{hmubarak@hbku.edu.qa}

\begin{document}

\onecolumn
\firstpage{1}

\title {Detecting and Reasoning of Deleted Tweets before they are Posted} 

\author[\firstAuthorLast ]{\Authors} 
\address{} 
\correspondance{} 

\extraAuth{}

\maketitle

\begin{abstract}
\section{}
Social media platforms empower us in several ways, from information dissemination to consumption. While these platforms are useful in promoting citizen journalism, public awareness etc., they have misuse potentials. Malicious users use them to disseminate hate-speech, offensive content, rumor etc. to gain social and political agendas or to harm individuals, entities and organizations. Often times, general users unconsciously share information without verifying it, or unintentionally post harmful messages. Some of such content often get deleted either by the platform  due to the violation of terms and policies, or users themselves for different reasons, e.g., regrets. There is a wide range of studies in characterizing, understanding and predicting deleted content. However, studies which aims to identify 
the fine-grained reasons (e.g., posts are offensive, hate speech or no identifiable reason) behind deleted content, are limited. In this study we address this gap, by identifying deleted tweets, particularly within the Arabic context, and labeling them with a corresponding fine-grained disinformation category. We then develop models that can predict the potentiality of tweets getting deleted, as well as the potential reasons behind deletion. Such models can help in moderating social media posts before even posting. 
\tiny
 \keyFont{ \section{Keywords:} Disinformation, Deleted Tweets, Twitter, keyword, keyword, keyword, keyword} 
\end{abstract}

\section{Introduction}

In the last decade, social media has become one of the predominant communication channels for freely sharing content online. 
The interactions on social media platforms enable public discussions online, such as those related to local issues and politics. Feelings of intolerance in media platforms usually generate and spread hate speech and offensive content through various communication channels. Such content can inflame tensions between different groups and ignite violence among their members. Malicious users intentionally and unintentionally use media platforms to impact people’s thoughts, disseminate hate speech, sway public opinions, attack the human subconscious, spread offensive content, fabricate truths, etc.
The misuse of social media platforms has turned them into potential grounds for sharing inappropriate posts, misinformation, and disinformation \citep{Zhou2016,alam-etal-2022-survey}. One type of inappropriate posts is \textbf{regrettable posts}, those that contain regrettable content, which can make the author feel guilty or can cause the intended audience to be harmed \citep{Zhou2016,sleeper2013read}. \textbf{Misinformation} is defined as ``\emph{unintentional mistakes such as inaccurate photo captions, dates, statistics, translations, or taking satire seriously}''. \textbf{Disinformation} is ``\emph{a fabricated or deliberately manipulated text/speech/visual context, and intentionally created conspiracy theories or rumors}''. While \textbf{melinformation} is ``\emph{defined as true information deliberately shared to cause harm}'' \citep{ireton2018journalism,alam-etal-2022-survey}.




\begin{figure}
\centering
	\includegraphics[width=.70\textwidth]{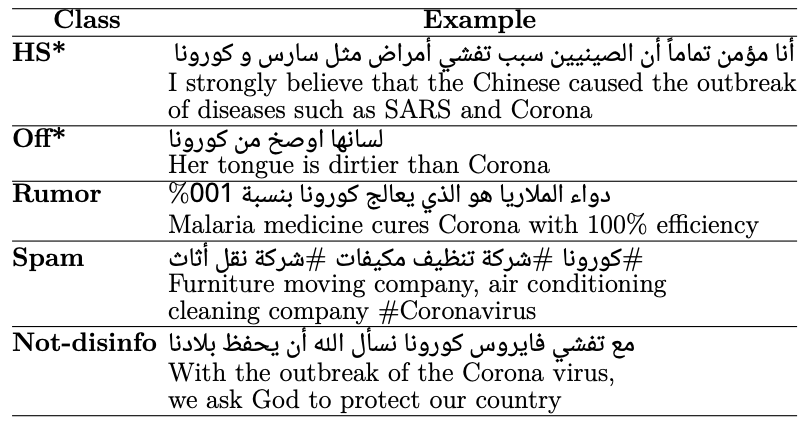} 
	\caption{Examples of disinformative and not-disinformative tweets. Not-disinfo: Not disinformative, HS: Hate speech, Off: Offensive.   
 \textbf{*WARNING:} Some examples have offensive language and hate speech, which may be disturbing to the reader
	}
	\label{fig:hs-targets}
\end{figure}

Such posts often get deleted for different reasons: 
{\em(i)} user themselves delete the posts, {\em(ii)} social media platform delete them due to breach of community guidelines \citep{Almuhimedi2013,wang2011regretted}. \citet{sleeper2013read} examined regrets within in-person and virtual conversations. They found that Twitter users tend to delete tweets or sometimes apologize once they realize their regret.  
The potential reasons behind tweets' deletion can be hate speech, offensive language, rumors, and/or spam that might violate community guidelines. In such cases, tweets get deleted, and users' accounts could get suspended as well. 
{\interfootnotelinepenalty10000 \footnote{\url{https://help.twitter.com/en/rules-and-policies/twitter-rules}} 
\footnote{\url{https://help.twitter.com/en/managing-your-account/suspended-twitter-accounts}}}
 
\citet{Bhattacharya_Ganguly_2021} stated that around 11\% of tweets are eventually deleted.
Although deleted tweets are not accessible once they are deleted, understanding the potential reasons behind their deletion motivates several researchers to understand and identify the content of regrettable tweets or tweets of suspended accounts  \citep{Zhou2016,wang2011regretted}. 
The importance of understanding the content of deleted tweets is the extraction of meaningful data of harmful content, and detecting and empowering users by sending warnings and suggestions before posts get shared on platforms.   
Prior studies have investigated detecting deleted tweets, spam accounts and their behaviors~\citep{stringhini2010detecting,lee2010uncovering}, analyzing regrets in bullying tweets \citep{xu2013examination}, and identifying factors for undesirable behavior such as spamming, negative sentiment, hate speech, and misinformation spread from deleted or suspended user accounts~\citep{toraman2022blacklivesmatter}. Most of such studies are limited to English language or distant supervision approach of labeling and fine-grained analysis. 

In this study, we investigate the following research questions:\\ \underline{{\em RQ1:}} What are the potential reasons (e.g., hate speech, offensive language) behind tweets' deletion?\\ \underline{{\em RQ2:}} Are deleted tweets a good way to collect different kinds of harmful content without imposing biases (ex: vs using keywords)?\\ \underline{{\em RQ3:}} How does Twitter deal with users who post disinformative content?\\ \underline{{\em RQ4:}} 
Can we detect potentiality of deletion of tweets and the corresponding reasons before they are posted?

To address these questions, we collected 40K deleted and non-deleted \textit{Arabic} tweets, and randomly selected a sample of 20K deleted and 2K non-deleted tweets. We then manually labeled them with fine-grained disinformative categories as shown in Figure \ref{fig:hs-targets} (See Section \ref{sec:dataset}). Using the labeled dataset, we trained models using classical algorithms (i.e., SVM, RF) and transformer that can detect the potentiality of  tweets getting deleted and the reasons of deletion. From our manual analysis, we found disinformative tweets with a proportion of 20\% and 7\% in deleted and non-deleted tweets, respectively. This clearly answers the question of deleted tweets being a good way to collect different kinds of harmful content, which can help in developing datasets and models to address disinformative content identification.
    
Our contributions and findings are summarized as the following:     
\begin{itemize}
    \item We developed a manually labeled dataset consisting of binary labels (deleted vs. non-deleted) and fine-grained disinformative categories. Our data collection method is generic and can be potentially applied to other languages and topics.
    \item Our proposed \textit{`detection and reasoning of deleted tweets'} approach can empower users by providing feedback before tweets are posted, which can also serve as a prevention mechanism while consciously and unconsciously producing and sharing disinformative posts. 
    \item Our data can be shared privately.
    \footnote{\textcolor{blue}{Note that we can only share text, like, share and annotated labels of the data, no information related to the user, which we deleted.}}
    \item We report insightful characteristics of deleted tweets' users by extracting their current activity status. 
    \item Our findings demonstrate that deleted tweets contain more disinformation than non-deleted ones. 
\end{itemize}
    
\section{Related Work}
Many research investigations have been conducted in the field of regretted and deleted social media data. However, what the literature lacks is the value deleted tweets could have if used as a source of data for essential NLP tasks such as disinformation detection. Starting with a set of 292K unique Twitter users, \citet{Almuhimedi2013} extracted all public tweets and/or retweets posted by users, as well as any replies to their posts alongside all relevant metadata of each tweet. Through the API, the authors could identify whether a tweet has been deleted, as ``a deletion notice was sent via the API containing identifiers for the user and the specific tweet'' \citep{Almuhimedi2013}. By doing so, they collected a total of 67.2M tweets, 65.6M of them were undeleted, and the other 1.6M were deleted. Through further analysis of the tweets, two of the reasons of deletion, the authors deemed as `superficial,' were due to typos and spam which made up 17\% and 1\% of the deleted tweets, respectively. 
Overall, the authors' analysis identified some common reasons of tweets' deletion. They also found that deleted and undeleted tweets share many common characteristics including the topics discussed within those tweets. 
Taking it a step further, \citet{Bhattacharya_Ganguly_2021} investigated the personality of users on Twitter by comparing users who deleted their tweets with the ones who did not. They started by randomly selecting 250K Twitter users and collected their corresponding tweets throughout August, 2015, as well as their corresponding deletion statuses.
 
Current literature suggests that deleted tweets are more likely to have aggressive and negative emotions.
\citet{Lugo2022} analyze `abusive' deletion behavior on Twitter. Using the Compliance Firehose Stream provided by Twitter, they extracted users who had more than 10 deletions over a one month period, which amounted to approximately 11 million users. They analyzed abusive deletion behaviour by extracting deletion volume, as well as frequency and life-span of deleted tweets. They found that `abusive' deleters tend to make use of this feature in order to manipulate the current 2,400 tweets a day limit set by Twitter. Other abusive deleters tend to continuously like and dislike a tweet in order to coordinate which tweets are to be more noticed by other users before deleting them. \citet{boyd2011}, on the other hand, focused on teenagers' deletion antics on social media. They suggested that teenagers tend to use deletion as a `structural' strategy to avoid receiving any judgements from their followers regarding any of their interests that they might express through social media posts.

Other researchers analyzed features and characteristics of deleted tweets with the goal of training models to predict the likelihood of deletion based on a number of features. \citet{Potash2016UsingTM} made use of topic modelling and word embeddings to predict whether a tweet is likely to be deleted or not, focusing on spam content. Using features such as tweet length, \# of links, ratio of upper-case text, hashtags, etc., they trained multiple classifiers, and tested them on a variety of datasets, resulting in a precision of approximately 81\%. Similarly, \citet{Bagdouri2015} investigated in the likelihood of a tweet gets deleted within 24hrs of its time posting. By analyzing features of both the deleted tweet, and the features of the corresponding users, they determined that tweets' features play a significant role in determining the likelihood of deletion. They specifically found that the device used to post the tweet is an important factor of determining deletion's potentiality. For instance, that tweets posted using smartphones were more likely to get deleted than those posted via computers. Furthermore, \citet{Gazizullina} utilized the Recurrent Neural Networks (RNN) to predict a tweet's likelihood of deletion using features about the text itself, as well as the metadata of tweets and users. Using post-processed word embeddings, they proposed a `Slingshot Net Model' which evaluated at an F-1 score of 0.755. 

Although there has been a good amount of researchers investigating deleted tweets and their characteristics, as far as we know, little work has been done in analyzing the role that disinformation plays in deleting tweets, specifically in the Arabic context. Therefore, we are inspired to contribute to the previous literature and conduct an investigation using Arabic deleted tweets to analyze the characteristics of deleted tweets, and identify different types of disinformation that could be shared within those tweets. 

\section{Dataset}
\label{sec:dataset}

\subsection{Data Collection}
\label{ssec:data_collection}

We used Twarc package\footnote{\url{https://github.com/DocNow/twarc}} to collect Arabic tweets having the word Corona in Arabic 
in February and March 2020. As mentioned in \citet{DBLP:journals/corr/abs-2201-06496}, this word is widely used by many people in all Arab countries, news media, and official organizations (e.g., the World Health Organization (WHO)) as opposed to the term COVID in Arabic.
The collection includes 18.8M tweets from which we took a random sample of 100K and checked their existence on Twitter in June 2022. We found that 64K tweets were still active, and 36K tweets were unavailable. The reasons of tweets' unavailability might be due to {\em(i)} users deleted tweets, {\em(ii)} accounts deleted, {\em(iii)} accounts suspended, or {\em(iv)} accounts became private. Note that accounts' deletion and suspension could also happen due to content violation of Twitter's policies. 

We selected a samples of tweets for the annotation in two phases, deleted and non-deleted tweets, respectively. In the \textit{first phase}, a random sample of 20K deleted tweets was selected for the manual annotation with fine-grained disinformative categories (see the following section). 
In the \textit{second phase}, we selected another 20K non-deleted tweets. From this set, we manually annotated a random sample of only 2K tweets with fine-grained disinformative categories. 
The reason of such two phases annotation from both deleted vs. non-deleted tweets was to see if there are similar proportions of disinformative categories in both sets. This also resulted to have an equal sample of 40K deleted and non-deleted tweets in which we used for the classification. 

\subsection{Annotation}
\label{ssec:annotation}
For the annotation, we selected major harmful categories (i.e., hate speech, offensive) discussed \citep{alam-etal-2022-survey,ijcai2022p781}.
Additionally, we selected rumor and spam categories as such content is posted on social media. Note that the intention behind rumors is not always harmful; however, due to the spread of false rumors on social media, they can turn into harmful content \citep{jungrumors2020}. According to Twitter policies,\footnote{\url{https://help.twitter.com/en/rules-and-policies/platform-manipulation}} these types of content are considered as platform manipulation content (``bulk, aggressive, or deceptive activity that misleads others and/or disrupts their experience''). 

We use the term ``disinformative'' to refer to \textit{hate speech (HS), offensive (Off), rumor and spam}. Worthy to be mentioned that not all categories directly fall under disinformation; however, we use this term to distinguish such categories from not-disinformative ones.

As for the annotation instructions, we follow the definition of these categories discussed in prior studies hate speech \citep{zampieri2020semeval}, offensive \citep{alam-etal-2022-survey,ijcai2022p781}, rumors \citep{jungrumors2020}, spam \citep{mubarak2020spam, RAO2021115742}. We asked annotators to select  \textit{not-disinformative} label if a tweet cannot be labeled as any of the disinformative categories we used in this study.

The annotation process consists of several iterations of training by an expert annotator, followed by final annotation. Given that tweets are in Arabic, we selected 
an Arabic fluent annotator of many Arabic dialects, with an educational qualification of Undergraduate and Master's degree. 

As mentioned earlier, in the \textit{first phase} we selected and manually annotated 20K deleted tweets. In the \textit{second phase}, we manually annotated 2K non-deleted tweets and rest of the 18K tweets of this phase are weakly labeled as \textit{not-disinformative}.

To ensure the quality of the annotations, during the first phase, two annotators annotated a randomly selected sample of 500 tweets (250  not-disinformative and 250 fine-grained disinformative tweets), then computed annotation agreement (see the next Section). 
Given that the annotation process is a costly procedure, we did not use more than one annotator for the rest of the tweet annotation.

\subsection{Annotation Agreement}
\label{ssec:annotation_agreement}
We assessed the quality of the annotations by computing inter-annotator agreement from the annotation of three annotators. We computed Fleiss $\kappa$ and average observed agreement (AoE)~\citep{fleiss2013statistical} which resulted in an agreement of 0.75 and 0.84, respectively. Based on the values, we reached \emph{substantial} agreement in the $\kappa$ measurement and \textit{perfect} agreement in the AoE measurement.\footnote{Note that, in the Kappa measurement, the values of ranges $0.41$-$0.60$, $0.61$-$0.80$, and $0.81$-$1$ refers to the moderate, substantial, and perfect agreement, respectively ~\citep{landis1977measurement}.}

\subsection{Statistics}
\label{ssec:statistics}

In Table \ref{tab:annotated_tweets_distribution}, we report the distribution of annotated tweets (deleted vs. non-deleted tweets). As mentioned earlier, for non-deleted tweets, we manually annotated 2K tweets, and the rest of them are weakly labeled as not-disinformative. From the table (phase 1 column), we observe that the distribution of disinformative tweets is relatively low compared to non-disinformative tweets, 19.7\%, and 80.3\%, respectively. From the given sample, 2K non-deleted manual annotated tweets (3rd column), we observe that the distribution between disinformative vs. non-disinformative tweets is 7.3\% and 92.7\%, respectively. Such a distribution clearly shows us that the distribution of disinformative tweets is more in deleted tweets than non-deleted tweets. This answers the first two questions (RQ1 and RQ2).

In the 4th column, we show the total number of tweets manually and weakly labeled from non-deleted tweets.

\begin{table}[]
\centering
\setlength{\tabcolsep}{2.5pt}
\scalebox{0.95}{
\begin{tabular}{@{}lrrr@{}}
\hline
\multicolumn{1}{c}{\textbf{Class label}} & \multicolumn{1}{c}{\textbf{\begin{tabular}[c]{@{}c@{}}Phase 1\\ Deleted\end{tabular}}} & \multicolumn{1}{c}{\textbf{\begin{tabular}[c]{@{}c@{}}Phase 2\\ Non-Deleted \\ (2K sample)\end{tabular}}} & \multicolumn{1}{c}{\textbf{\begin{tabular}[c]{@{}c@{}}Phase 2\\ Non-Deleted\end{tabular}}} \\ \hline
Not-Disinfo & 16,066 & 1,854 & 19,854 \\
HS & 2,180 & 58 & 58 \\
Off & 735 & 47 & 47 \\
Rumor & 252 & 29 & 29 \\
Spam & 767 & 12 & 12 \\\hline
Total & 20,000 & 2,000 & 20,000 \\ \hline
\end{tabular}
}
\vspace{-3mm}
\caption{Distribution of annotated tweets.}
\vspace{-4mm}
\label{tab:annotated_tweets_distribution}
\end{table}

\section{Analysis}

\label{sec:analysis}

We present an in-depth analysis of the deleted tweets dataset to gain a better understanding of the topics and entities being tweeted about, in relation to COVID-19, and the users who authored those tweets. This includes identifying {\em(i)} most common rumors discussed about COVID-19 within this dataset; {\em(ii)} the most common hate-speech targets within the dataset; 
{\em(iii)} the current activity status of the users to analyze the potential role that could have been played in the deletion of their tweets; and other metadata such as the distribution of different attributes (e.g., hashtags, user mentions) and, retweet and follower counts. 

\subsection{Rumors}
\label{ssec:rumors}

When doing the manual annotation, we kept track the frequent rumors based on the semantic meaning.\footnote{There are no duplicate tweets; we removed them at the beginning.} The most common rumors were regarding finding potential cures and/or medication to battle COVID-19, while other rumors are related to conspiracies regarding the long-term effects of COVID-19 on humans, as well as potential preventative measures to minimize the spread of the virus. In table \ref{tab:rumor_examples}, we list the most frequent rumors shared by users included within the dataset, ordered by descending order of frequency. 

\begin{table}[h]
\centering
\setlength{\tabcolsep}{2.5pt}
\scalebox{0.95}{
\begin{tabular}{@{}l@{}}
\hline
\multicolumn{1}{c}{\textbf{Examples}} \\ \hline
\begin{tabular}[c]{@{}l@{}}\textbf{1.} A number of drugs, including Malaria, Influenza, \\ and AIDS drugs help coronavirus patients improve.\end{tabular} \\
\textbf{2.} Coronavirus is an American invention. \\
\begin{tabular}[c]{@{}l@{}}\textbf{3.} Coronavirus is a biological warfare weapon, \\ and many people and novels predicted the virus ahead of time.\end{tabular} \\
\begin{tabular}[c]{@{}l@{}}\textbf{4.} Coronavirus damages organs of the human body \\ such as the brain and genitals as it causes male infertility.\end{tabular} \\
\begin{tabular}[c]{@{}l@{}}\textbf{5.} Having certain foods such as tea, maamoul and gum\\ prevents the infection of Coronavirus.\end{tabular} \\
\begin{tabular}[c]{@{}l@{}}\textbf{6.} Religious rituals such as wearing niqab, burning incense, \\being Muslim, and ablution prevents the infection.\end{tabular} \\ \hline
\end{tabular}
}
\vspace{-2mm}
\caption{Most frequent rumors. Translated forms of Arabic tweets.}
\vspace{-4mm}
\label{tab:rumor_examples}
\end{table}

\subsection{Hate Speech Targets}
\label{ssec:hate_speech_targets}
We wanted to understand if hate speech is targeted toward any entities, countries, or organizations. During the manual annotation, we identified targets to which hate speech has been targeted. We then identified 
the most frequent entities mentioned throughout tweets classified as hate speech. Countries, political parties, and religion seem to be the most common entities found in tweets that include hate speech words/phrases. In Figure \ref{fig:hs-targets}, we report most frequent hate speech targets. 

\begin{figure}
\begin{center}
	\includegraphics[width=.40\textwidth]{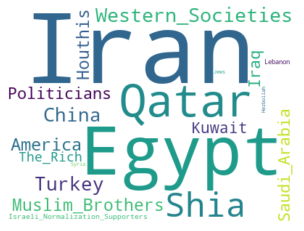} 
	\vspace{-2mm}
	\caption{Word cloud for most frequent hate-speech targeted topics/entities. 
	}
	\vspace{-4mm}
	\label{fig:hs-targets}
\end{center}
\end{figure}

\subsection{User Status}
\label{ssec:user_status}
We wanted to understand if there are any association of disinformative categories and current Twitter users' status. The goal was to understand whether the current status of a given account is a major factor of deleting tweets. Whereas if the account gets deleted or suspended, tweets of such account get deleted as well. 
Using the information provided by Twitter API, we determined the current user status of all unique users who posted at least one disinformative tweet. 
In total, there were 3,677 unique users who posted at least one disinformative tweet. 
Each of the unique users was classified under one of four categories: suspended (removed by Twitter), deleted (initiated by the user), active-private (user is active but private, blocking public access to any of their tweets), and active-public (user is active, and their tweets are publicly available). 

\begin{figure}[ht]
\begin{center}
	\includegraphics[width=.49\textwidth]{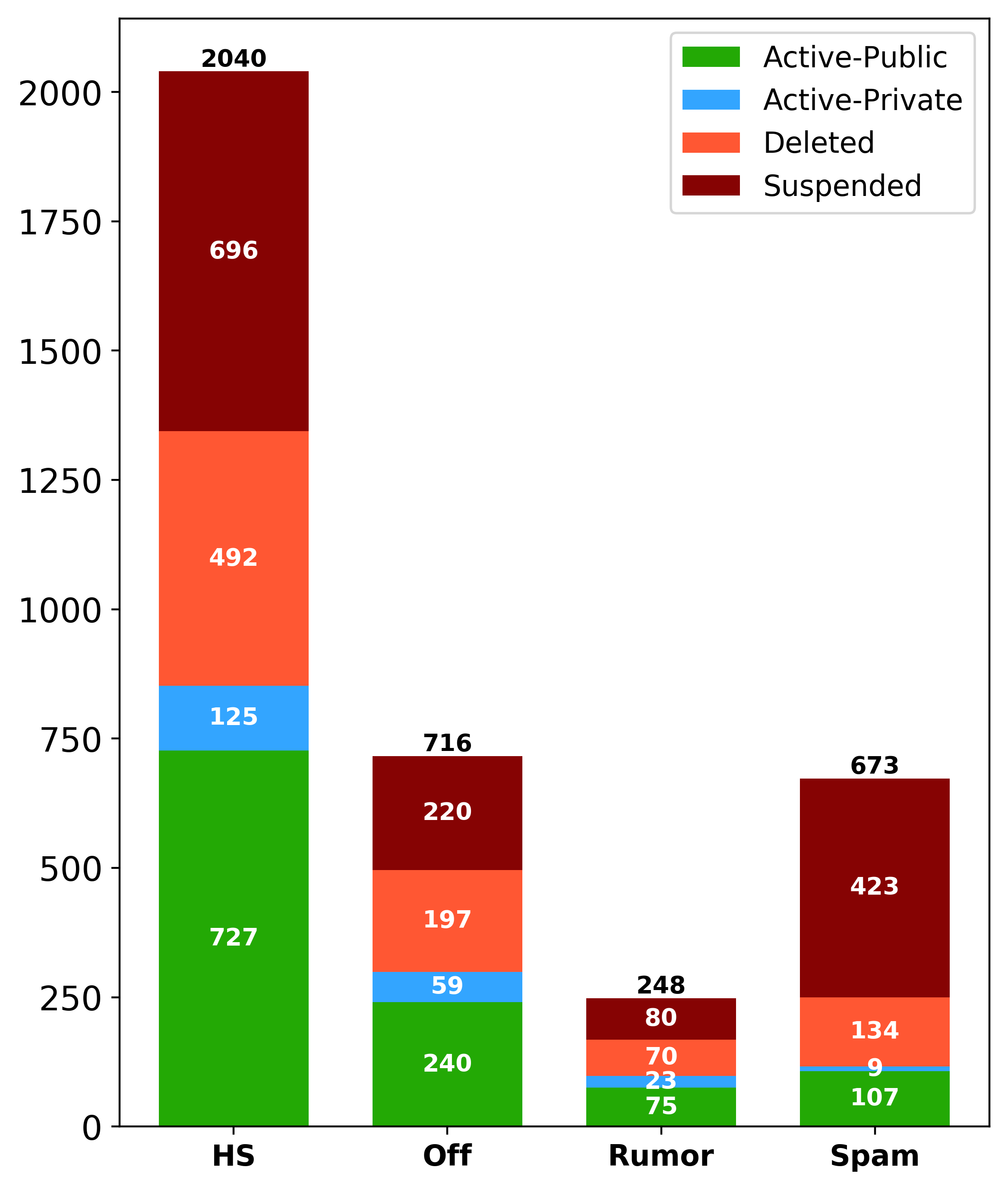} 
	\vspace{-3mm}
	\caption{Distribution of users' account status corresponding to each disinformative category.
	This status is based on the time of our analysis period (August, 2022)}
	\vspace{-2mm}
	\label{fig:userstatus}
\end{center}
\end{figure}

In Figure \ref{fig:userstatus}, we present the number of users' accounts for each disinformative categories. From the figure, we observe that the distribution of hate speech is higher than other categories.
An interesting point to note is that almost 40\% (1,419) of all users, with at least one disinformative post, were suspended by Twitter. Out of those users, Twitter was very efficient at identifying and disabling spam users, as it could suspend 423 accounts of users who shared at least one spam tweet, which amounts to more than 62\% of accounts that posted any spam content. In respect of hate speech posters, Twitter identified and suspended over 34\% (696) of them. For, the other accounts, approximately 24\% (893) of them were deleted by the users themselves, while 6\% (216) of them are currently active but are set to private, and the remaining 33\% (1,224) are still active and public. This analysis answers RQ3, as it shows that Twitter is able to identify some users who post disinformative content, and ultimately suspend the whole account. 

As a result, user status is an important factor to take into consideration when analyzing and characterizing the deletion of tweets, as it could be due to their corresponding accounts that are not existing anymore, either as a result of Twitter suspension, user deactivation, or the user setting the account to private. 

\subsection{Other Metadata}
\label{ssec:other_metadata}
In Table \ref{tab:del_nodel_stats}, we report the distributions of some attributes in the non-deleted, deleted, and the associated disinformative tweets. There are minor differences between the non-deleted and disinformative tweets. However, the subset of the deleted tweets that is labeled as disinformative has different distributions. For example, disinformative tweets have double as many URLs, as well as more replies than the other sets, and they are less likely to be retweeted by one seventh (12\% vs 77\% or 82\%).

From this dataset, we also observe that the percentage of hate speech is higher than other categories, which might be due to the topic of interest, i.e., COVID-19. Similar findings are reported in \citet{mubarak2020arcorona}, which suggest that tweets about COVID-19 were found to have higher percentage of hate speech (7\%) as it's a polarized topic, e.g., attacking some countries for spreading the virus. This is typically different than random collections of Arabic tweets. \citet{mubarak2020arabic} reported that the percentage of offensive language in random collections is between 1\% to 2\%, and hate speech ratio is even less. 


We hypothesize that many of the deleted tweets contain more harmful content than normal (ex: 10.9\% hate speech, 3.8\% spam), and Twitter deleted them as they violate its community standards 
or they were deleted by the users themselves as they regretted posting some tweets because they contain offensiveness or rumors. This also answers our first two research questions. 

\begin{table}[!htb]
\centering
\scalebox{0.9}{
\begin{tabular}{lrrr}
\hline

\multicolumn{1}{c}{\bf{Attributes}} & \multicolumn{1}{c}{\begin{tabular}[c]{@{}l@{}}\textbf{Non-Deleted}\\ \end{tabular}} & \multicolumn{1}{c}{\begin{tabular}[c]{@{}l@{}}\textbf{Deleted} \end{tabular}} & \multicolumn{1}{c}{\begin{tabular}[c]{@{}l@{}}\textbf{Disinformative} \end{tabular}}\\ \hline
\bf{Hashtags} & 57\% & 55\% & \bf 63\%\\ 

\bf{URLs} & 29\% & 25\% & \bf 51\%\\

\bf{User Mentions} & 82\% & \bf 87\%& 24\%\\ 

\bf{Replies} & 05\%& 05\% &\bf 09\%\\

\bf{Retweets} & 77\% & \bf 82\% & 12\%\\ \hline
\end{tabular}}
\caption{Percentages of tweets having different attributes.}
\label{tab:del_nodel_stats}
\end{table}

\section{Experiments and Results}
\label{sec:experiments}
\begin{figure}
\begin{center}
	\includegraphics[width=.43\textwidth]{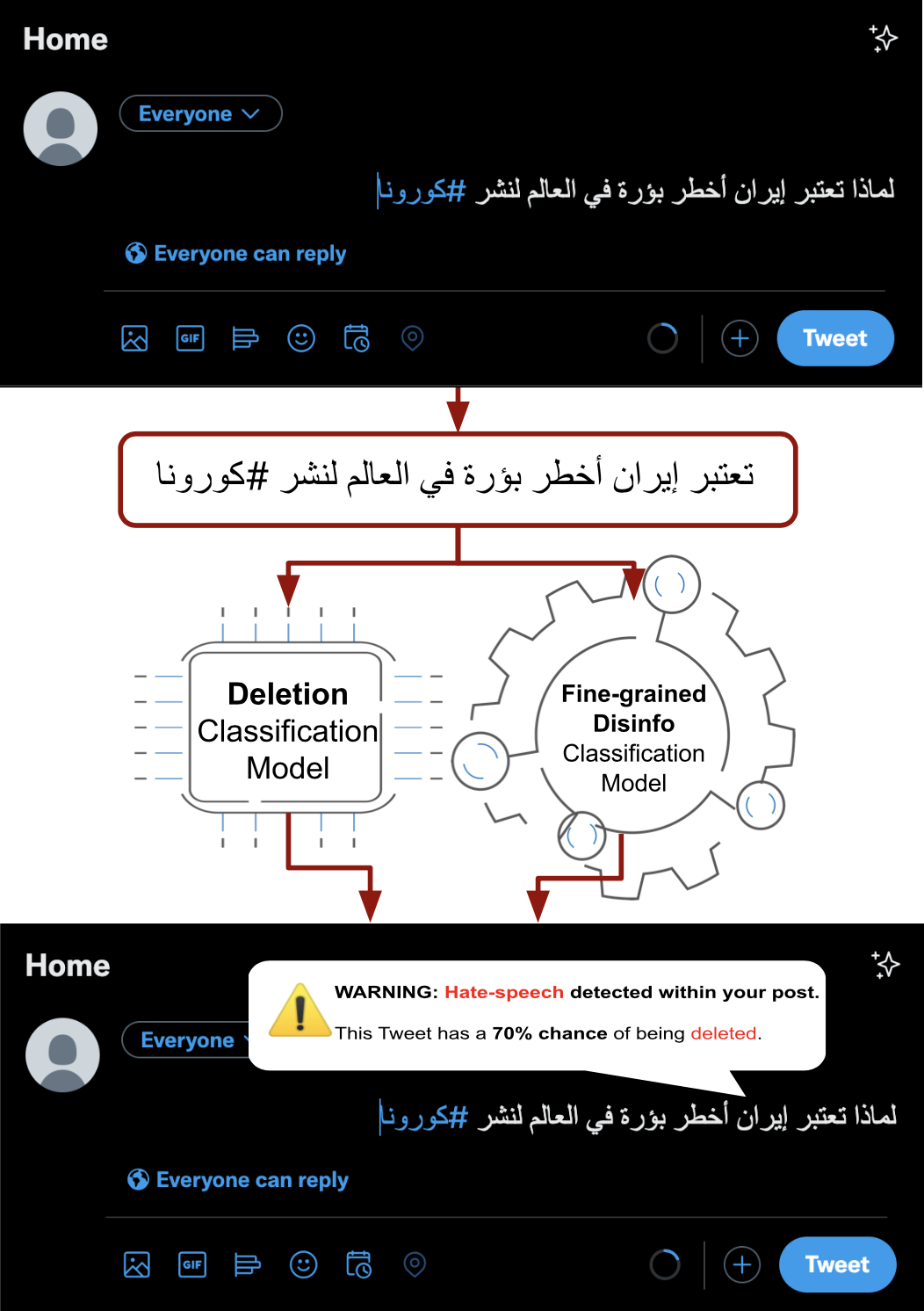} 
	\caption{A pipeline of our proposed system to detect and warn users while posting -- what can happen and why. \underline{\textbf{Translation (HS*):}} \textit{Why is Iran considered the most dangerous spot in the world for spreading Corona?}}
	\vspace{-4mm}
	\label{fig:system-figure}
\end{center}
\end{figure}

In Figure \ref{fig:system-figure}, we present our proposed pipeline of post deletion detection with reasons while posting on social media. While posting the tweet detection model can detect whether a tweet will be deleted, then fine-grained disinformation model can detect whether it is one of the disinformation categories (e.g., in this case, hate speech). Our goal is to empower users while posting and/or sharing content and reduce the spread of misleading and harmful content. In the following sections, we describe the details of the proposed models and results.

\subsection{Experiment Settings}
\label{ssec:exp_settings}
We have conducted different classification experiments with a focus on detecting whether a tweet can be deleted before posting, and what could be the possible reasons. We train three different classifiers as follows: {\em(i)} a binary classifier to detect whether a tweet will be deleted using the labels \textit{deleted} vs. non-deleted tweets, which consists of 40K tweets; 
{\em(ii)} a binary classifier to detect whether a tweet disinformative vs not-disinformative (binary classification setting) {\em(iii)} a multiclass classifier to detect fine-grained disinformative categories. For the latter two classifiers we used manually labeled 22K tweets. Note that we have not used all 40K for the later two sets of experiments given that they have weakly labeled (18K considered as not-disinformative) tweets. This could be a part of our future study.

\subsection{Data Splits and Preprocessing}
\label{ssec:data_splits}
To conduct experiments, we split our dataset into three subsets with a 70-10-20 setting for train, dev and test sets, respectively. 
The class distributions within each subset are shown in Table \ref{tab:class_dist_sets}. The second set (ii) of data split in the Table is a subset of the first set, whereas the third set (iii) is only fine-grained \textit{Disinformation} categories of the second set (ii). 

\begin{table}[]
\centering
\setlength{\tabcolsep}{8.0pt}
\scalebox{1.0}{
\begin{tabular}{@{}lrrrr@{}}
\hline
\multicolumn{1}{c}{\textbf{Class label}} & \multicolumn{1}{c}{\textbf{Train}} & \multicolumn{1}{c}{\textbf{Dev}} & \multicolumn{1}{c}{\textbf{Test}} & \multicolumn{1}{c}{\textbf{Total}} \\ \hline
\multicolumn{5}{c}{\textbf{(i) Binary: Deleted vs. Non-deleted}} \\ \hline
Deleted & 14,012 & 2,020 & 3,968 & 20,000 \\
Not-deleted & 13,988 & 1,980 & 4,032 & 20,000 \\ \hline
\textbf{Total} & \textbf{28,000} & \textbf{4,000} & \textbf{8,000} & \textbf{40,000} \\ \hline
\multicolumn{5}{c}{\textbf{(ii) Binary: Disinfo vs. Non-disinfo}} \\ \hline
Disinformation & 2,879 & 394 & 807 & 4,080 \\
Not-Disinfo & 12,521 & 1,806 & 3,593 & 17,920 \\ \hline
\textbf{Total} & \multicolumn{1}{l}{\textbf{15,400}} &  \multicolumn{1}{l}{\textbf{2,200}} & \multicolumn{1}{l}{\textbf{4,400}} & \multicolumn{1}{l}{\textbf{22,000}} \\ \hline
\multicolumn{5}{c}{\textbf{(iii) Multiclass: Fine-grained disinfo labels}} \\ \hline
HS & 1,563 & 227 & 448 & 2,238 \\
Off & 554 & 83 & 161 & 798 \\
Rumor & 189 & 31 & 61 & 281 \\
Spam & 550 & 67 & 146 & 763 \\ \hline
\textbf{Total} & \textbf{2,856} & \textbf{408} & \textbf{816} & \textbf{4,080} \\ \hline
\end{tabular}
}
\vspace{-2mm}
\caption{Distribution of the dataset for different experimental settings for train, dev and test sets.}
\vspace{-4mm}
\label{tab:class_dist_sets}
\end{table}

\subsubsection{Preprocessing:} 
Given that social media texts are normally noisy. Before any classification experiments, we applied preprocessing to the dataset. The preprocessing includes the removal of hash symbols and non-alphanumeric symbols, URL replacement with a ``URL'' token, and username replacement with ``USER'' token.

\subsection{Models}
We experimented with binary and multiclass settings both classical and deep learning algorithms discussed below. The classical models include {\em (i)} Random Forest (RF) ~\citep{breiman2001random}, and {\em (ii)} Support Vector Machines (SVM) ~\citep{platt1998sequential}, which was most widely reported in the literature. The other reason to choose such algorithms is that they are computationally efficient and useful in many production systems. 

Given that large-scale pretrained Transformer models have achieved state-of-the-art performance for several NLP tasks. Therefore, as deep learning algorithms, we used deep contextualized text representations based on such pretrained transformer models. We used AraBERT \citep{baly2020arabert} and multilingual transformers such as XLM-R~\citep{conneau2019unsupervised}. For Transformer models, we used the Transformer toolkit~\citep{Wolf2019HuggingFacesTS}. We fine-tuned each model using the default settings for ten epochs as described in \citet{devlin2018bert}. We performed ten reruns for each experiment using different random seeds, and selected the model that performed best on the development set. 

\subsection{Results}
We report accuracy (Acc),  weighted precision (P), recall (R), and F1 scores which take into account class imbalance that we had in our dataset. We compute majority as a baseline.

In Table \ref{tab:exp_results}, we report the classification experiments of all different settings. From the table, we can see that all models outperform the majority class baseline. Comparing to the classical algorithms, SVM outperforms RF in two settings out of three. While comparing monolingual vs multilingual transformer models, we observe that AraBERT performs well in detecting deleted tweets, XLM-R outperforms well in classifying whether the text of the tweet is disinformative or not. For classifying fine-grained disinformative categories, AraBERT outperforms all other models. Our results clearly answer \textit{RQ4}, in that we can detect potentiality of deletion of tweets and the corresponding reasons, with reasonable accuracy. 

\begin{table}[]
\centering
\setlength{\tabcolsep}{13.0pt}
\scalebox{1.0}{
\begin{tabular}{@{}lrrrr@{}}
\hline
\multicolumn{1}{c}{\textbf{Model}} & \multicolumn{1}{c}{\textbf{Acc}} & \multicolumn{1}{c}{\textbf{P}} & \multicolumn{1}{c}{\textbf{R}} & \multicolumn{1}{c}{\textbf{F1}} \\ \hline
\multicolumn{5}{c}{\textbf{(i) Binary: Deleted vs. Non-deleted}} \\ \hline
Majority & 0.496 & 0.246 & 0.496 & 0.329 \\
RF & 0.896 & 0.882 & 0.896 & 0.854 \\
SVM & 0.852 & 0.851 & 0.852 & 0.850 \\
AraBERT & 0.910 & 0.896 & 0.910 & \textbf{0.902} \\
XLM-R & 0.886 & 0.784 & 0.886 & 0.832 \\  \hline
\multicolumn{5}{c}{\textbf{(ii) Binary: Disinfo vs. Non-disinfo}} \\  \hline
Majority & 0.817 & 0.667 & 0.817 & 0.734 \\
RF & 0.853 & 0.871 & 0.853 & 0.812 \\
SVM & 0.837 & 0.838 & 0.837 & 0.837 \\
AraBERT & 0.888 & 0.882 & 0.888 & 0.884 \\
XLM-R & 0.897 & 0.894 & 0.897 & \textbf{0.895} \\  \hline
\multicolumn{5}{c}{\textbf{(iii) Multiclass: Fine-grained disinfo labels}} \\  \hline
Majority & 0.537 & 0.288 & 0.537 & 0.375 \\
RF & 0.696 & 0.760 & 0.696 & 0.622 \\
SVM & 0.669 & 0.677 & 0.669 & 0.665 \\
AraBERT & 0.755 & 0.757 & 0.755 & \textbf{0.752} \\
XLM-R & 0.762 & 0.747 & 0.762 & 0.745 \\ \hline
\end{tabular}
}
\vspace{-3mm}
\caption{Classification results for different settings that can detect tweet deletion and possible fine-grained reasons. XLM-R: XLM-RoBERTa}
\vspace{-4mm}
\label{tab:exp_results}
\end{table}

\subsubsection{Error Analysis}
We analyzed all rumors and offensive tweets that are misclassified as hate speech (n=243). We found annotation errors in 18\% of the cases, and 5\% of the errors are due to sarcasm, negation or tweets having rumors and hate speech in the same time. In the other cases, the model predicted the label as hate speech as it is the dominant class as shown in statistics in Table \ref{tab:annotated_tweets_distribution}. By looking into individual class label performance for disinformative categories, we observe that spam  and hate speech are the best-performing labels (F1=0.940 and F1=0.779, respectively). The offensive label is the lowest in performance (F1=0.513), which is due to the mislabeling as hate speech in many cases.

\section{Conclusion and Future Work}
\label{sec:conclutions}
We presented a large manual annotated dataset that consists of deleted and non-deleted Arabic tweets with fine-grained disinformative categories. We proposed classification models that can help in detecting whether a tweet will be deleted before even being posted and detect the possible reasons of the deletion. We also reported the common characteristics of the users whose tweets were deleted.Our findings suggest that deleted tweets can be used in developing annotated datasets of misinformative and disinformative categories.
Future work will include more fine-grained categories which are mostly harmful (e.g., racist) and find more reasons of tweets' deletion which can empower social media users. In addition, we plan to explore multitask learning setup that can reduce computational cost and may boost the performance of the model. Also, for future explorations regarding this topic, there needs to be a larger dataset of deleted tweets used that takes into consideration factors such as the account being suspended as opposed to the individual tweet being deleted.

\section{Limitations}
We developed a dataset that consists of tweets extracted from Twitter only. Additionally, we developed models that require an exploration to understand whether models will work on datasets from other social media platforms. 

It is important to note that although this exploration looks into the likelihood of tweet deletion based on an annotated dataset, the moderation techniques employed by social media networks such as Twitter require further analysis to be able to gain insight into potential reasons for user suspension and/or tweet deletion. 

\section*{Ethical Consideration and Broader Impact}
Our dataset was collected from Twitter, following its terms of service. It can enable an analysis of social media content that may be an area of interest to social media platforms and users. Our models can help to reduce the intentional and unintentional posting of social media posts that can mislead and/or harm social media users. 

For reproducibility concerns, we aim to share the dataset privately that may limit to widely access the dataset. However, we are looking into ethical issues if even privately sharing them is allowed.


\bibliographystyle{Frontiers-Harvard} 
\bibliography{bibliography}

\end{document}